# The Carbon Footprint of Machine Learning Training Will Plateau, Then Shrink

David Patterson[1,2], Joseph Gonzalez[2], Urs Hölzle[1], Quoc Le[1], Chen Liang[1], Lluis-Miquel Munguia[1], Daniel Rothchild[2], David So[1], Maud Texier[1], and Jeff Dean[1]

***Abstract***: Machine Learning (ML) workloads have rapidly grown in importance, but raised concerns about their carbon footprint. Four best practices can reduce ML training energy by up to 100x and $CO_2$ emissions up to 1000x. By following best practices, overall ML energy use (across research, development, and production) held steady at <15% of Google's total energy use for the past three years. If the whole ML field were to adopt best practices, total carbon emissions from training would reduce. Hence, we recommend that ML papers include emissions explicitly to foster competition on more than just model quality. Estimates of emissions in papers that omitted them have been off 100x–100,000x, so publishing emissions has the added benefit of ensuring accurate accounting. Given the importance of climate change, we must get the numbers right to make certain that we work on its biggest challenges.

**Keywords/Index terms**: I.2.6.g Machine learning < I.2.6 Learning < I.2 Artificial Intelligence < I Computing Methodologies, I.2.7 Natural Language Processing < I.2 Artificial Intelligence < I Computing Methodologies, B.9 Power Management < B Hardware, C.1.3.i Neural nets < C.1.3 Other Architecture Styles < C.1 Processor Architectures < C Computer Systems Organization, C.4 Performance of Systems < C Computer Systems Organization

## 1. Introduction

Over the past few years, a growing number of papers have highlighted the carbon emissions of machine learning (ML) workloads. While this work has been instrumental in rightfully elevating the discussion around carbon emissions in ML, some studies significantly overestimated actual emissions, which in turn led to worrisome extrapolations [1,2]:

> *The answers are grim: Training such a model would cost US$100 billion and would produce as much carbon emissions as New York City does in a month.*

Recent work highlights the complexities and nuances associated with carbon accounting for ML and more broadly computing workloads [3, 4, 5]. In this paper, we

- Describe four practices that reduce the energy usage and carbon emissions of ML workloads by orders of magnitude relative to traditional choices;
- Show that these practices help keep ML under 15% of Google's total energy use for the past three years; and
- Explain why published faulty estimates and extrapolations are 100x–100,000x higher than the real carbon footprints.

Responsible AI is a broad topic; we focus on a single issue that has received much attention from the ML community and public: carbon emissions from ML training. Emissions can be classified as

- *Operational*, the energy cost of operating the ML hardware including datacenter overheads, or

- *Lifecycle,* which additionally includes the embedded carbon emitted during the manufacturing of all components involved, from chips to datacenter buildings.

Like prior work we focus on operational emissions; estimating lifecycle emissions is a larger, future study.

We identified best practices that can reduce energy by up to 100x and carbon emissions by up to 1000x compared to following orthodox choices ("4Ms"):

---

[1] Google
[2] University of California, Berkeley

1. *Model.* Selecting efficient ML model architectures while advancing ML quality, such as sparse models versus dense modes, can reduce computation by factors of ~5–10.

2. *Machine.* Using processors optimized for ML training such as TPUs or recent GPUs (e.g., V100 or A100), versus general-purpose processors, can improve performance/Watt by factors of 2–5.

3. *Mechanization.* Computing in the Cloud rather than on premise improves datacenter energy efficiency[3], reducing energy costs by a factor of 1.4–2.

4. *Map.* Moreover, Cloud computing lets ML practitioners pick the location with the cleanest energy[4], further reducing the gross carbon footprint by factors of 5–10[5].

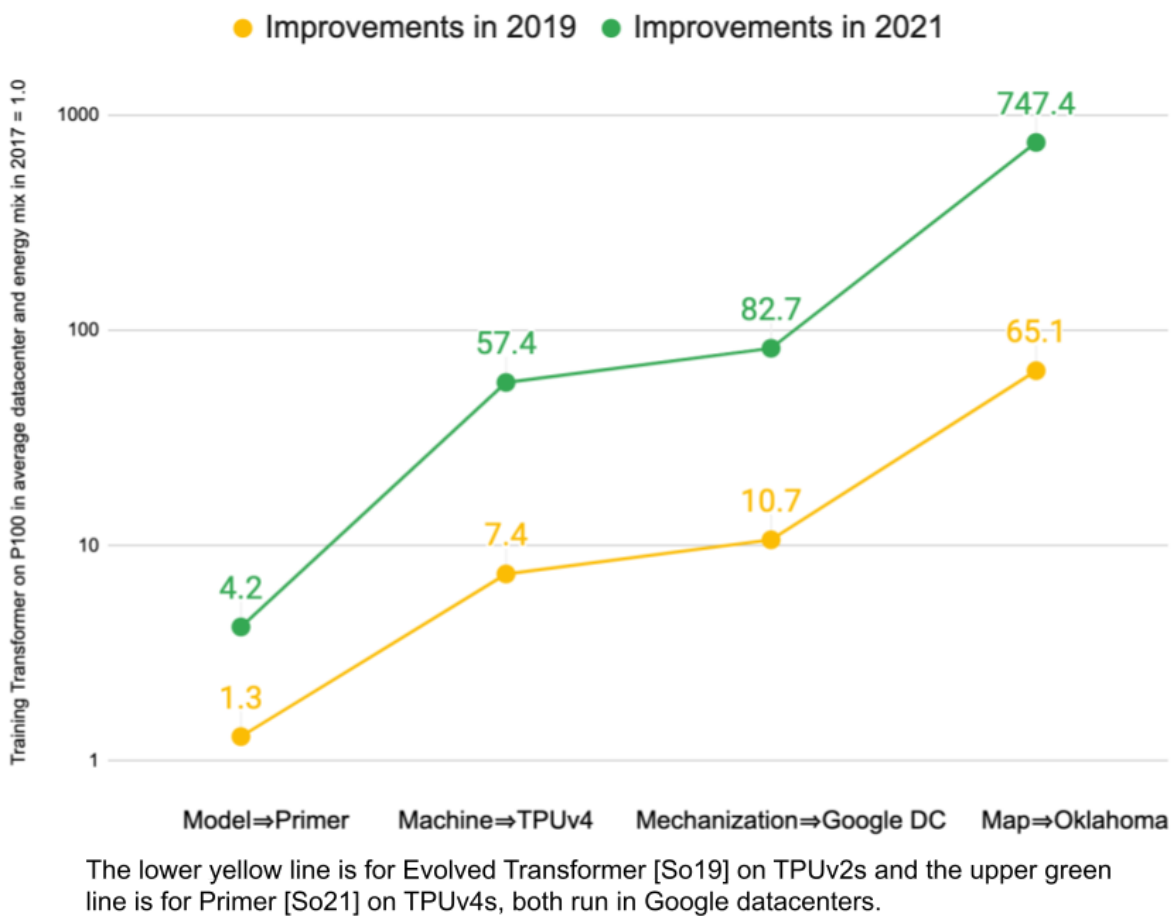

The lower yellow line is for Evolved Transformer [So19] on TPUv2s and the upper green line is for Primer [So21] on TPUv4s, both run in Google datacenters.

**Figure 1. Reduction in gross $CO_2$ emissions since 2017 from applying best practices (Section 3). They show large end-to-end improvements, broken down into the 4Ms. Gross $CO_2$ emissions here *excludes* Google's carbon neutral and 100% renewable energy credits, and reflect Google's 24/7 CFE methodology [5].**

Figure 1 illustrates how four good choices together reduce energy consumption by 83x and $CO_2$ emissions by 747x over four years while maintaining the same quality. The original modeled estimate represents training the Transformer model in 2017 on an ML-oblivious GPU[6] in an average datacenter using an average energy mix (like [2]). The yellow line shows optimizations possible in 2019, the green line optimizations possible today. In both cases, optimized ML hardware reduces energy consumption significantly, with newest-generation hardware (TPUv4) providing an additional 2.4x over the 2019 hardware (TPUv2). Using efficient cloud datacenters and a low-carbon datacenter region per Google's 24/7 carbon-free energy (CFE) methodology further reduces the carbon footprint by another order of magnitude (note the log-scale Y axis), resulting in a 747-fold reduction in carbon footprint compared to the original estimate. In this paper, *gross $CO_2$ emissions* are the carbon emissions resulting from a workload in a particular location before any compensating actions.

---

[3] The Cloud uses custom warehouses designed for energy efficiency, whereas on-premise datacenters are inefficiently located in smaller, older spaces intended for other purposes. HPC datacenters are efficient, but can't enable shifting to green locations.
[4] Most data transmission power is for the network equipment of the Internet even when idle [3]. In comparison, shipping photons over fiber optics is relatively trivial.
[5] Using a carbon-neutral cloud like Facebook or Google further reduces the footprint to zero because they match 100% of operational energy use with renewable energy. We exclude those offsets.

Supported by the results in Figure 1 and in Section 3, we predict that if ML communities embrace these best practices, the carbon footprint of ML training will shrink over this decade.

Below we summarize this paper's contributions:

- Two studies show the impact of best practices: a 750x emissions reduction without loss of accuracy from Transformer (Figure 1) and a 14x emissions reduction from GPT-3 by the larger GLaM model that improves accuracy.

- Location choice, even within one country, can significantly impact the carbon footprint.

- We provide the first report by a hyperscaler company of the percentage of its overall energy use devoted to ML training and inference.

- We show that the carbon footprint of searching for better ML models can reduce the impact of downstream ML tasks by much more than the cost of the search.

- We describe how following best practices significantly reduced the energy consumption and carbon footprint of training compared to the faulty estimates commonly cited [2,6,7].

## 2. Overview of Energy and $CO_2$e for ML Training

We estimate energy and carbon footprints using these terms:

- *$CO_2$ equivalent emissions* (*$CO_2$e*) accounts for carbon dioxide ($CO_2$) and all the other greenhouse gasses as well: methane, nitrous oxide, and so on.

- *Metric tons* are the common $CO_2$e unit of measure, abbreviated as *$tCO_2$e*, representing 1000 kilograms (2205 pounds).

- *Megawatt hours (MWh)* measure energy; one MWh equals 1,000,000 Watts of electricity used continuously for 1 hour. One *Terawatt hour* (TWh) equals 1,000,000 MWh.

- [Power Usage Effectiveness](#) (*PUE*) is the industry standard metric of datacenter efficiency, defined as the ratio between total energy usage (including all overheads, like cooling) divided by the energy directly consumed by the datacenter's computing equipment. The average industry datacenter PUE in 2020 was [1.58](#) (58% overhead) while cloud providers have PUEs of ~1.10 [5].

- *Carbon intensity* ($tCO_2$e per MWh) is a measure of the cleanliness of a datacenter's energy. The average datacenter carbon emissions in 2020 was [0.429](#) $tCO_2$e per MWh but the gross $CO_2$e per MWh can be 5x lower in some Google datacenters.

---

[6] The 2016 NVIDIA P100 was optimized for graphics, not ML.

The energy consumption of the servers performing a training task is proportional to the number of processors used and the duration of the training run:

$$MWh\ =\ Hours\ to\ train\ \times\ Number\ of\ Processors\ \times\ Average\ Power\ per\ Processor$$

We include all server components in "Processor" (including local memory, network links, and so on). Additionally, the datacenter consumes energy to power and cool the hardware (e.g., voltage transformation losses, cooling equipment), captured by PUE. Thus, the final formula for energy consumption:

$$MWh\ =\ (Hours\ to\ train\ \times\ Number\ of\ Processors\ \times\ Average\ Power\ per\ Processor)\ \times\ PUE$$

We can then turn energy into carbon by multiplying it with the carbon intensity of the energy supply:

$$tCO2e\ =\ MWh\ \times\ t\ CO2e\ per\ MWh$$

The real-world values for many factors are readily available. ML practitioners usually publish the number and type of processors and hours to train, and the power consumption of most hardware components is well known or can be measured accurately. Many Cloud companies publish the PUE of their datacenters.

In comparison, carbon intensity is harder to obtain. For this paper we use the carbon intensity of Google datacenters, derived from Figure 2. We hope other providers will publish so that carbon intensity can be compared across datacenters.

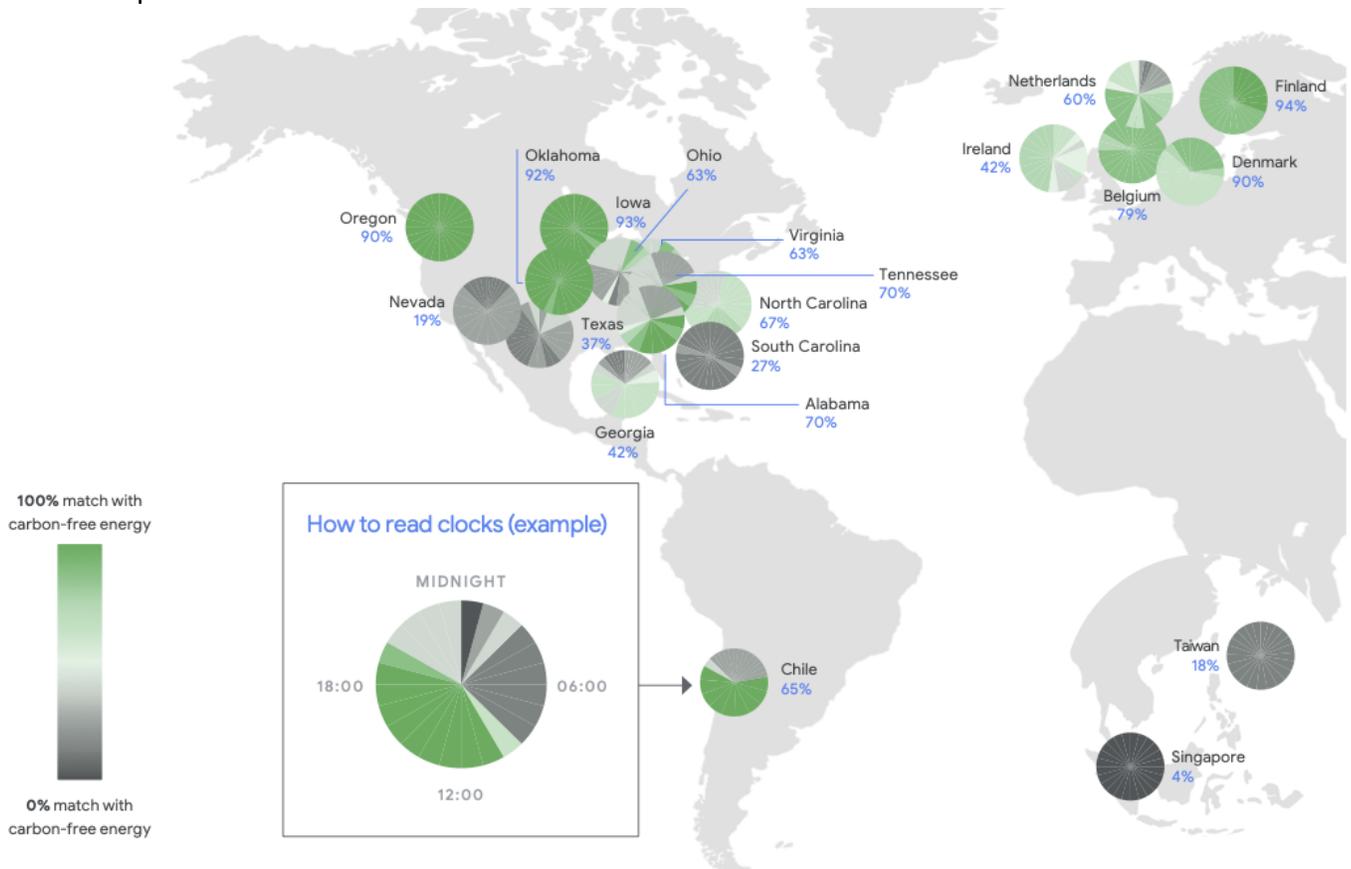

**Figure 2. Percent Carbon Free Energy by Google Cloud Location in 2020. The map shows the %CFE and how the percentage changes by time of day. Chile has a high %CFE from 6AM to 8PM, but not at night. The US examples on this map range from 19% CFE in Nevada to 93% in Iowa, which has strong prevailing winds both night and day. (sustainability.google/progress/energy/)**

# 3. Case Study 1: Transformer vs. Evolved Transformer vs. Primer

Many of the headline grabbing advances in AI stem from *deep neural networks* (*DNNs*); indeed, three DNN leaders shared the 2018 ACM A.M. Turing Award.

DNN computations have two phases: *training*, which constructs accurate models through an intensive computational process involving iterative updating of *parameters*, and *inference*, which uses the trained models to generate outputs from model inputs. ML practitioners use different models for different tasks: object recognition, language translation, and so on. Training "learns" parameters that raise the likelihood of correctly mapping from input to result. Unlike in traditional computing, the actual DNN code is relatively small. The "smarts" come from training DNNs from millions of labeled examples versus writing millions of lines of code.

The *Transformer* model debuted in 2017 and is used primarily for natural language processing (NLP). Its distinguishing feature is focusing attention on portions of its input. Two years later, So et al used *neural architecture search* (*NAS*) to discover the *Evolved Transformer* model that matched Transformer's quality scores but was ~1.3x faster [8]. In 2021, a different NAS found the Primer model that again matched the quality scores but was 4.2x faster than the original Transformer [9].

Figure 1 above plots the end-to-end reduction in $CO_2e$ by applying the best practices from Section 1. The reference point is the Transformer model trained on a P100 GPU in an average on-premise datacenter with the average PUE of 1.60 in 2017 and using the average $tCO_2e$ per MWh of 0.488. Here are the practices (4Ms):

1. *Model.* In 2019 the best model was Evolved Transformer, in 2021 Primer.

2. *Machine.* Compared to the unoptimized P100s from 2017, the ML-optimized TPUv2 in 2019 and TPUv4 in 2021 reduce energy consumption by 5.7x and 13.7x, respectively. This reduction is a function of both improved logic (more specialized hardware), newer chip fabrication technology, and more efficient mapping of the training task to hardware (better utilization of the functional units) [10].

3. *Mechanization.* The third point shows a reduction of 1.4x from the better PUE of Google's Cloud datacenter versus the average datacenter.

4. *Map.* A big surprise was how much location of the datacenter affected carbon intensity (Figure 2). In 2019, the datacenter in the US region with the highest CFE score was Oklahoma with a score of 96%, and in 2020 it was Iowa at 93%.

To summarize, following best practices yields a 65x reduction in $CO_2e$ two years after Transformer was introduced. Two additional years later—with ML model, hardware, and energy mix improvements—another 11x was possible, for an overall reduction of 747x. These drastic overall improvements, as well as their trajectory over time, suggest that extrapolating current parameters to predict future $CO_2e$ is fraught with peril.

# 4. Case Study 2: GPT-3 vs GLaM

Next is a large NLP model that received considerable attention in the ML community and in the press in 2020: *GPT-3* is an autoregressive language model with 175B parameters, 10x more than any non-sparse language model at the time, and 100-1000x more than most other ML models [11]. To put GPT-3 into perspective, its predecessor GPT-2 had 1.5B parameters, and the Transformer models above used ≤0.2B. Developed by OpenAI, GPT-3 was trained on 10,000 V100 GPUs[7] in a Microsoft cloud datacenter.

---

[7] The 2017 NVIDIA V100 *is* optimized for ML.

A winner of the best paper award at NeurIPS, the recent GPT-3 paper already has >2500 citations and made mainstream media headlines. One benefit of large models like GPT-3 is that they don't need to be retrained for every new task—called few-shot generalization—unlike smaller models.

GLaM is a new language model using 7x more parameters than GPT-3. It is a mixture of experts model that only activates experts selectively based on the input so that no more than 95B parameters (8%) are active per input token. The dense GPT-3 activates all 175B parameters on every token. More parameters and sparsity allow GLaM to exceed GPT-3 on quality *and* efficiency [12].

Figure 3 compares them. GPT-3 took 405 V100 years to train in 2020. OpenAI trained in the Microsoft cloud to leverage a low PUE but with an energy mix that matched the US datacenter average [5]. In comparison, GLaM trained on TPUv4s in 2.8x fewer accelerator years, using 2.8x less energy than GPT-3. Additionally, GLaM ran in the Oklahoma datacenter where the $tCO_2e$ per MWh was ~5x lower (0.088 vs 0.429). Evolved Transformer and Primer improve energy use and $CO_2e$ while maintaining quality scores, but GLaM improves all three metrics.

ML researchers are continuously improving the efficiency of large language models through innovations in algorithms and model architectures. Only 18 months after GPT-3, GLaM can reduce the gross carbon footprint by ~14x despite raising accuracy. These drastic improvements again show that extrapolating current ML trends to predict future ML energy use and $CO_2e$ can greatly overestimate consumption, as there are continuous, significant improvements in algorithms and hardware.

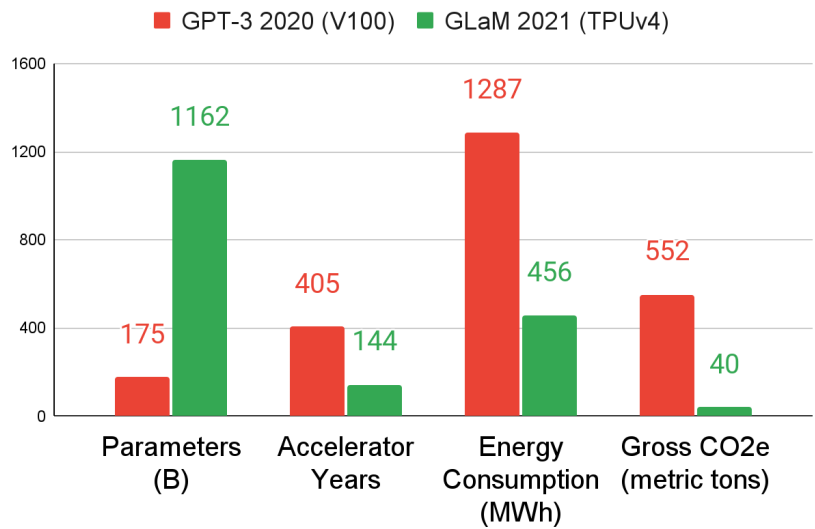

**Figure 3. Parameters, accelerator years of computation, energy consumption, and gross $CO_2e$ for GPT-3 and GLaM. If instead of outperforming GPT-3 on quality scores, GLaM was only trained to match, it would halve the time, energy, and $CO_2e$. Google's renewable energy purchases further reduce the impact to zero.**

## 5. Overall ML Energy Consumption

The previous sections investigate the energy consumption of a single training task. Here we discuss the overall footprint of all ML workloads at a major user of ML, Google.

Many hyperscalers regularly publish their energy consumption metrics. According to their sustainability reports, the annual energy consumption in 2020 was 15.4 TWh for Google and 10.8 TWh for Microsoft. These reports put the training energy of large models into perspective. Training GPT-3 was ~0.012% of Microsoft's energy consumption in 2020 and GlaM was ~0.004% of Google's. For further comparison, the portion of the 22,000 people from 68 countries who in 2019 flew to attend the two major ML conferences (NeurIPS and CVPR) collectively had a $CO_2e$ impact arguably had ~10x–100x higher than the impact of training of all the ML models in this paper [5].

While Google's overall energy consumption increases as usage grows, our data shows that despite the growth of ML applications, the ML *portion* of Google's overall energy consumption is not increasing. To estimate that fraction, we measured the energy consumption (including datacenter overheads) of the following components:

- All TPUs and GPUs in Google datacenters, including associated dedicated servers and networking equipment. Virtually all ML training executes on TPUs and GPUs and most inference as well. We can differentiate training versus inference runs on TPUs and GPUs.

- Any CPU consumption attributable to ML inference (no significant training was done solely on CPUs).

To estimate the CPU portion of inference, we inspected [Google-Wide Profiling](#) results to measure CPU consumption of the libraries used for ML inference. We then converted CPU utilization into energy consumption using sensors that measure server power. Our numbers likely overestimate because some libraries are used in non-ML cases as well. Also, we may double count some host CPUs already accounted for in the TPU/GPU measurements, and some GPU use is for graphics.

We retroactively performed these calculations based on data for one week of April in 2019, 2020, and 2021. Each time the ML portion was 10% to 15% of Google's total energy consumption for that week despite ML representing 70%-80% of the FLOPS at Google[8]. While ML usage certainly increased during these three years, algorithmic and hardware improvements kept that growth to a rate comparable with overall energy growth at Google. Across all three years, about ⅗ of ML energy use is for inference and ⅖ for training. These measurements include all ML energy usage: research, development, testing, and production.

Consequently, we take the stable fraction for ML as a strong indication that despite ML's increasing popularity, if we follow best practices its energy consumption is not skyrocketing, contrary to commonly expressed fears. This stability may reflect economic factors in addition to technical factors: after all, everything has a budget, and budget limits encourage efficient use of ML resources.

Worldwide datacenter energy consumption [is not growing quickly either](#). [4] observes that global datacenter energy consumption increased by only 6% from 2010 to 2018, despite datacenter computing capacity increasing by 550% over the same time period, and contrary to 2010 predictions of a 70% increase by 2018. One key factor was the shift from conventional datacenters to cloud datacenters. Not only are cloud datacenters often more efficient, cloud servers typically have significantly higher utilization than on-premise servers. That allows the same workloads to be served with less hardware and thus less energy, just as books purchased for libraries are more frequently read than books purchased for home use. As of 2021 only 15%-20% of all workloads have moved to the Cloud [13], so there is still plenty of headroom for Cloud growth to replace inefficient on-premise datacenters.

Finally, most cloud companies compensate at least partially for their carbon emissions. In particular, Google (since 2017) and Facebook (since 2020) purchase enough renewable energy annually to match 100% of their usage, so each MWh of new consumption is offset by one MWh of new renewable energy, albeit not necessarily in the same location. Microsoft's similar goal is for 2025. Thus, the net carbon impact of ML computations for some companies could be considered zero. Such multi-billion dollar direct energy purchases by hyperscalers have spurred the growth of renewable energy substantially: in some countries, they are more significant investors in renewable energy than government subsidies [14].

## 6. Additional Factors

For completeness, we will briefly address two other concerns about ML energy usage: the impact of Neural Architecture Search (NAS), which may run thousands of training runs as part of a single

---

[8] See Section 7 for an explanation why using many more FLOPS does not imply using much more energy.

search—potentially exploding overall energy consumption—and ML's impact on client-side energy usage.

A commonly expressed concern is that automated methods might increase training energy consumption. As the name implies, NAS uses computers to find models with higher quality or efficiency than human experts can find. NAS is generally not performed once per model training, but once per *problem domain+architectural search space combination.* Evolved Transformer and Primer are examples of the benefits of NAS [8, 9]. NAS has also been applied to find models that have better quality *and* run faster by adapting them to a given processor [15].

The NAS producing Evolved Transformer used 7.5 MWh. The use of Evolved Transformer while training the large Meena model saved 15x the energy cost of this NAS [5]. Finding the even faster Primer used only 6.2 MWh. Overall, NAS is a net environmental gain if the discovered model is trained more than a few times. Often, the more efficient models found by NAS are open-sourced and reused hundreds or even thousands of times [5]. Consequently, as a whole, it is likely that NAS reduces total ML energy usage by producing more efficient models whose downstream use more than compensates for the initial search effort.

To estimate ML energy use on client devices, [16] studied mobile phones. Most modern phones have ML accelerators; for example, the Google Pixel 6 phone has an Edge TPU, which runs most of the ML workload. During a typical day, the Edge TPU consumes less than 1% of the phone's energy. Client use of ML libraries and ML applications (bar code reading, OCR, face recognition, etc) played a similar minor role. CPU and GPU account for ~8% of total energy usage on phones, of which a small fraction is for ML. To be on the safe side, we use a generous upper bound for ML energy usage on today's mobile phones of 5%.

The estimated global energy use of the 3.8B mobile phones in 2021 is ~7.9 TWh, assuming nightly charging and accounting for charger inefficiency [16]. The upper bound for ML on mobile phones is then 0.4 TWh. Google's ML server energy use in 2020 was ~6 times higher than this conservative estimate of ML on all mobile phones. This calculation does not include the other energy consumption of ML at other cloud companies, so server-side ML energy usage clearly dominates client-side usage.

## 7. Related Work

[17] is a similar study that also provides a framework to understand the potential climate impacts of ML research. It also offers a leaderboard to foster competitions on reducing $CO_2e$ of ML and a tool to collect energy use and $CO_2e$ from the preliminary training runs. [5] is a 22-page technical report by the authors of this paper that goes into greater detail on the many of the issues here.

[18] warns of the danger of "Red AI", which focuses on model quality gains regardless of the training cost or $CO_2e$. They encourage embracing "Green AI" where the focus is on computing efficiency as well as model quality. Arguing that it can be difficult to measure energy and $CO_2e$, they recommend minimizing the number of floating point operations (FLOPs) to train a model. Alas, FLOPs is not a good metric, for time and energy can be uncorrelated with FLOPs. For example, AutoML found faster models that used 2.4x as many FLOPs [15]. An underlying reason is that main memory accesses are much slower and use much more energy than FLOPs today. A DRAM access is ~6000x energy of a 16b FLOP (1300 vs 0.21 picoJoules) [10]. Another reason is that scaling up FLOPS/sec is much easier for ML accelerators than scaling up memory bandwidth[9]. To improve efficiency further, ML practitioners should focus more on reducing memory accesses than on FLOPs. More successful attempts to simplify calculation of energy are online calculators, such as the ML Emissions Calculator [5,19].

---

[9] These ratios and the ease for hardware to scale up FLOPS/second also helps explain how ML energy use can be under 15% despite ML being responsible for 70%–80% of the FLOPS over the past three years.

The opening quote in Section 1 is based on a 2019 study from the University of Massachusetts (UMass) that estimated the environmental impact of training [2]. More than 1000 papers cite this paper as the source for the impact on carbon emissions of ML models, e.g., [1,6,7,17,18,19,20]. The study calculated the energy consumed and carbon footprint of the NAS by [8] that led to Evolved Transformer. The UMass estimate was 284 $tCO_2e$ for NAS; the actual number was only 3.2 $tCO_2e$, a factor of 88 smaller. The reasons for the overshoot:

1. Since the authors of the original NAS paper didn't include energy and emissions for Google systems, the UMass estimate was based on older GPUs not optimized for ML[10] instead of TPUv2 and on the average datacenter PUE and average carbon intensity instead of the real numbers for a Google datacenter. This difference explains 5x.

2. There was also confusion about the computational cost of NAS. Described subtly in [8], the Evolved Transformer NAS used a small proxy task to search for the best models to save time, money, and energy, and then scaled up the found models to full size. However, [2] assumed the search was done with full size tasks. The resulting computation estimate for NAS was another 18.7x too high

The actual overshoot was 18.7x for computation and 5x for Google versus the average datacenter, so the real emissions for the one-time search were 88x less (3.2 versus 284 tCO2e).

The faulty estimates in [2] are understandable given the lack of access to internal information. It is likewise understandable that those estimates were propagated in other papers, like [1,6,7,17,18,19,20]. Unfortunately, some papers that cite this work confused the one-time cost of the NAS of [8] with the relatively tiny "every-time" cost that is incurred from training. This cost difference is more than 1000x[11]. This confusion led them to believe Evolved Transformer used more than 2 million GPU hours to train, cost millions of dollars, and its emissions were five times the lifetime of a car (284,019 kg) [6,7]. In reality, training the medium Evolved Transformer, which achieves the same accuracy level as the Transformer-big model, takes 120 TPUv2 hours, costs $40, and emits only 2.4 kg (0.00004 car lifetimes), 120,000x less. This gap is nearly as large as if one overestimated the $CO_2e$ to *manufacture* a car by 100x and then used that number as the $CO_2e$ for *driving* a car.[12]

Accuracy is difficult if estimated retrospectively, as evidenced by the difference between these published estimates and actual measurements. This example underlines the importance of our recommendation that *authors* calculate and publish energy consumption and carbon footprint.

## 8. Conclusion

Machine Learning (ML) workloads have rapidly grown in importance, raising legitimate concerns about their energy usage. Fortunately, the real-world energy usage trend of ML is fairly boring. While overall energy use at Google grows annually with greater usage, the percentage for ML has held steady for the past three years, representing <15% of total energy usage. Inference represents about ⅗ of total ML energy usage at Google, owing to the many billion-user services that use ML. GLaM, the largest natural language model trained in 2021, improved model quality yet produced 14x less $CO_2e$ than training the previous state-of-the art model from 2020 (GPT-3) and used only 0.004% of Google's annual energy.

Furthermore, we illustrated that in large scale production ML deployments, minimizing emissions from training is not the ultimate goal. Instead, the combined emissions of training and serving need to be minimized. Approaches like neural architecture search increase emissions but lead to more efficient serving and a strong overall reduction of the carbon footprint of ML. Another perspective is that some consider the carbon footprint to be erased entirely if the cloud provider matches 100% of their energy consumption with renewable energy, as Google and Facebook have done and as Microsoft will soon do.

---

[10] They used the P100. The most recent GPU available was the V100, which is much faster in part because it was optimized for ML, unlike the P100.
[11] The NASs for Evolved Transformer and Primer produce 1347x and 1618x more $CO_2e$, respectively, than their training.
[12] The [average US car trip](…) produces [4 kg](…), but manufacturing a car generates [9200 kg of CO2e](…) (2300x more).

While ML workloads have grown rapidly over the past decade, and while the number of computations per training run has similarly increased by orders of magnitude, our data shows that technology improvements have largely compensated for this increased load. We believe this consistent overall low percentage is a testimony to the benefits of following best practices:

- *Datacenter providers* should publish the PUE, %CFE, and $CO_2$e/MWh per location so that customers who care can understand and reduce their energy consumption and carbon footprint.

- *ML practitioners* should train using the most effective processors in the greenest datacenter that they have access to, which today is often in the Cloud.

- *ML researchers* should continue to develop more efficient ML models [8,9], such as by leveraging sparsity [12] or by integrating retrieval into a smaller model. They should also publish their energy consumption and carbon footprint, both in order to foster competition on more than just model quality and to ensure accurate accounting of their work, which is difficult to do accurately post-hoc.

These numbers may vary across companies, but the practices we've identified are applicable to virtually all ML training workloads and open to all to use. As a result, we predict that if all ML communities embrace these best practices, we can create a virtuous circle that will bend the curve so that in this decade we'll see the total carbon footprint of ML training at first plateau and then shrink.

Finally, we show that published studies overestimated the cost and carbon footprint of ML training because they didn't have access to the right information or because they extrapolated point-in-time data without accounting for algorithmic or hardware improvements.

Climate change is important, so we must get the numbers right to ensure that we work on the biggest challenges. Many efforts are underway to reduce the operational energy and $CO_2$e of ML training, as illustrated by the 4Ms: model, machine, mechanization, and map. Thus, within information technology, we believe the biggest climate change challenge is not the operational cost of ML but more likely the lifecycle cost of manufacturing computing equipment of all types and sizes[13].

**Acknowledgement**. We had a great deal of help from others along the way for an earlier study [5] that eventually led to this version of the paper. Emma Strubell made several suggestions for the prior paper, including the recommendation to examine the recent giant NLP models. Christopher Berner, Ilya Sutskever, OpenAI, and Microsoft shared information about GPT-3. Dmitry Lepikhin and Zongwei Zhou did a great deal of work to measure the performance and power of GPUs and TPUs in Google data centers. Hallie Cramer, Anna Escuer, Elke Michlmayr, Kelli Wright, and Nick Zakrasek helped with the data and policies for energy and $CO^2$e emissions at Google. Talia Ringer provided helpful suggestions on how to better present related work.

## 9. References

[1] Thompson, N.C., et al., 2021. Deep Learning's Diminishing Returns: The Cost of Improvement is Becoming Unsustainable. *IEEE Spectrum.*

[2] Strubell, E., et al., 2019. Energy and policy considerations for deep learning in NLP. *Annual Meeting of the Association for Computational Linguistics*.

[3] Koomey, J., and Masanet, E., 2021. Does not compute: Avoiding pitfalls assessing the Internet's energy and carbon impacts. *Joule*, 5(7), pp.1625-1628.

[4] Masanet, E., et al., 2020. Recalibrating global datacenter energy-use estimates. *Science*, 367(6481).

---

[13] IT manufacturing for 2021 included 1700M cell phones, 340M PCs, and 12M data center servers.

[5] Patterson, D., et al., 2021. Carbon Emissions and Large Neural Network Training. arxiv:2104.10350.

[6] Thompson, N.C., et al., 2020. The computational limits of deep learning. arxiv:2007.05558.

[7] Freitag, C., et al., 2021. The real climate and transformative impact of ICT: A critique of estimates, trends, and regulations. *Patterns*, 2(9).

[8] So, D.R., et al., 2019. The Evolved Transformer. International Conference on Machine Learning.

[9] So, D.R., et al., 2021. Primer: Searching for efficient transformers for language modeling. Conference on Neural Information Processing Systems.

[10] Jouppi, N., et al., 2021, Ten Lessons From Three Generations Shaped Google's TPUv4i, International Symposium on Computer Architecture.

[11] Brown, T.B., et al., 2020. Language models are few-shot learners. Conference on Neural Information Processing Systems.

[12] Du, N., et al., 2021. GLaM: Efficient Scaling of Language Models with Mixture-of-Experts. arxiv:2112.06905.

[13] Evans, B. 2021, Amazon Shocker: CEO Jassy Says Cloud Less than 5% of All IT Spending, https://cloudwars.co/amazon/amazon-shocker-ceo-jassy-cloud-less-than-5-percent-it-spending/

[14] Schechner, S., 6/23/2021, Amazon and Other Tech Giants Race to Buy Up Renewable Energy, *Wall Street Journal.*

[15] Li, S., et al., 2021 Searching for Fast Model Families on Datacenter Accelerators, Conference on Computer Vision and Pattern Recognition.

[16] Patterson, D., et al., 2022, Estimating ML Energy and Carbon Footprint in Smartphones, in preparation.

[17] Henderson, P., et al., 2020. Towards the systematic reporting of the energy and carbon footprints of machine learning. *Journal of Machine Learning Research*.

[18] Schwartz, R., et al., 2020. Green AI. *Communications of the ACM*, 63(12).

[19] Lacoste, A., et al., 2019. Quantifying the carbon emissions of machine learning. arxiv:1910.09700.

[20] Bender, E.M., et al., 2021, On the Dangers of Stochastic Parrots: Can Language Models Be Too Big?. *ACM Conference on Fairness, Accountability, and Transparency.*

## Short Author Bios

David Patterson is a Distinguished Engineer in the Google Brain project in Mountain View, California, 94043 USA, the vice-chair of the RISC-V Foundation's board of directors, Director of the RISC-V International Open Source Laboratory, and a professor emeritus at the University of California, Berkeley. His research interests include domain-specific computer architectures and open instruction set architectures. He has a PhD in computer science from the University of California, Los Angeles. Contact him at pattrsn@berkeley.edu.

Joseph Gonzalez is a professor of Computer Science at the University of California, Berkeley, California 94720. His primary research interests are in the design of systems for machine learning as well as efficient neural architectures. He has worked on projects ranging from large-scale language modeling and efficient computer vision to the design of systems for graphical model inference, real-time prediction serving, and autonomous driving. He has a PhD in Machine Learning from Carnegie Mellon University. Contact him at jegonzal@berkeley.edu.

Urs Hölzle is the senior vice president of operations at Google and a Google Fellow. His interests include large scale clusters, cluster networking, Internet performance, and datacenter design. Hölzle received a PhD in computer science from Stanford University. Contact him at urs@gmail.com.

Quoc Le is a principal scientist at Google Brain. His research interests include AI, AutoML, natural language understanding and computer vision. He has a PhD in Computer Science from Stanford University and Bachelor at The Australian National University. Contact him at [qvl@google.com](qvl@google.com)

Chen Liang is a researcher in Google Brain. His works focus on the integration of machine learning and symbolic representations. His research interests include automated machine learning (AutoML), neural symbolic methods, natural language understanding and program synthesis. He has a PhD in Computer Science from Northwestern University and a B.S. in Physics from Peking University. Contact him at [crazydonkey200@gmail.com](crazydonkey200@gmail.com).

Lluis-Miquel Munguia is a Senior Software Engineer at Google, where he works on co-design for deep-learning accelerators. His main research interests include the performance analysis of special-purpose computer architectures, power efficiency, and high performance computing. He has a PhD in computational science and engineering from the Georgia Institute of Technology. Contact him at [llmunguia@google.com](llmunguia@google.com).

Daniel Rothchild is a PhD student at the University of California, Berkeley advised by Joseph Gonzalez. Research interests include distributed and federated learning, and machine learning for drug discovery and materials design. Contact him at [drothchild@berkeley.edu.](drothchild@berkeley.edu.)

David R. So is a Senior Research Engineer in the Google Brain project. His research focuses on language modeling, AutoML, and improving deep learning efficiency. Contact him at [davidso@google.com](davidso@google.com).

Maud Texier is Head of Energy Development at Google. She leads a team responsible for developing and scaling 24/7 carbon-free energy for Google's data centers. Her research interests are in carbon abatements technologies, carbon-free energy technologies, grid systems modernization and decarbonization. Maud holds an MS of engineering in Energy and Power systems from Ecole Centrale Paris. Contact her at [maudt@google.com](maudt@google.com).

Jeff Dean is Google Senior Fellow and the Senior Vice President of Research at Google, where he co-founded the Google Brain project, and has worked on a variety of machine learning and software systems that underlie Google's products, many of which are open-sourced. His research interests include large-scale distributed systems, machine learning, applications of machine learning, information retrieval, microprocessor architecture, and compiler optimizations. He has a PhD in computer science from the University of Washington. Contact him at [jeff@google.com](jeff@google.com).